\documentclass{article}

\usepackage[utf8]{inputenc}
\usepackage[T1]{fontenc}
\usepackage{natbib}
\usepackage{amssymb}
\usepackage{amsmath}
\usepackage{xcolor}
\usepackage{bbm}
\usepackage{url}
\usepackage{booktabs}
\usepackage{multirow}
\usepackage{caption}
\usepackage{graphicx}
\usepackage{float}
\usepackage{bm}
\usepackage{authblk}

\begin{document}

\title{Attribution Modeling Increases Efficiency of Bidding in Display Advertising}

\newcommand*\samethanks[1][\value{footnote}]{\footnotemark[#1]}

\author[1]{Eustache Diemert\thanks{contributed equally - contact \{e.diemert,j.meynet\}@criteo.com}}
\author[1]{Julien Meynet\samethanks[1]}
\author[2]{Pierre Galland}
\author[3]{Damien Lefortier\thanks{work done while at Criteo}}
\affil[1]{Criteo Research}
\affil[2]{Criteo}
\affil[3]{Facebook}

\renewcommand\Authands{ and }

\begin{abstract}
Predicting click and conversion probabilities when bidding on ad exchanges is at the core of the programmatic advertising industry. Two separated lines of previous works respectively address i) the prediction of user conversion probability and ii) the attribution of these conversions to advertising events (such as clicks) after the fact.
We argue that attribution modeling improves the efficiency of the bidding policy in the context of performance advertising.

Firstly we explain the inefficiency of the standard bidding policy with respect to attribution. Secondly we learn and utilize an attribution model in the bidder itself and show how it modifies the average bid after a click. Finally we produce evidence of the effectiveness of the proposed method on both offline and online experiments with data spanning several weeks of real traffic from Criteo, a leader in performance advertising.
\end{abstract}

\maketitle

\section{Introduction}
Performance advertising has become a very successful model of programmatic advertising where advertisers payment is based on delivered value as measured by events of interest. Two main models exist: cost-per-click (CPC) or cost-per-action (CPA). We focus on the latter where the advertiser attributes conversion events (e.g. a signup or more generally a sale) to one (or more) advertising events such as displays or clicks. The industry standard for conversion attribution is to credit only the last click in the 30 days before the conversion although this is changing recently \cite{Ji9781}.

For advertising agencies and platforms the state of the art strategy is the Expected Value Bidder (EVB) that bids the expected value of the opportunity \cite{Perlich2012} \cite{Rosales2012}. EVB has been proven to be a dominant strategy in the case of non-repeated second-price auctions \cite{Easley2010}. The expected value is computed as advertiser payment times the predicted conversion probability. Numerous previous studies focused on accurately estimating click or conversion probabilities at scale \cite{Chapelle2014} \cite{Rosales2012} and open datasets were published for this task \citep{Criteo2014} \cite{Zhang2014}. 

On the other side, conversion attribution has been thoroughly studied in the fields of game theory and econometrics \cite{Shao2011} \cite{Abhishek2012b} \cite{Dalessandro2012} and mechanisms have been proposed to better match the advertiser payment with the ad effectiveness, especially when multiple advertising events or channels are involved.

In order to improve the EVB policy one can research the usage of attribution prediction. A motivational example is to consider consecutive auctions for the same user. If the user has clicked after the first impression the advertising platform has good chances of getting the attribution, in the sense that a consecutive conversion would be attributed to that click - unless another advertising event captures the attribution in between. In a subsequent auction, a rational bidding policy should take into account this non-zero attribution probability and lower its bid compared to the first one as there is little chance that a second click would increase the attribution probability. The EVB policy is not able to act rationally in that respect since it doesn't keep track nor predicts the attribution probability evolution. As a first exploratory step towards improved real-time bidding policies for advertising we propose to leverage an attribution model to modify the bidding strategy.

We posit that an attribution bidding policy has the benefit of better aligning the incentives of the advertiser with the ones of the advertising platform and the user, resulting in increased efficiency for all parties involved. Intuitively this comes from a better representation of the effect of advertising over time: the advertiser should pay for ads that affect attributed conversions, the platform should reduce its costs when bidding for ads with low attribution probability and the user should see less ads once engaged with the advertiser.

The rest of the paper is organized as follows. Firstly we describe our core contribution: a suitable attribution model in Section \ref{ssec:attrmodel} and the related modifications to the bidding strategy in Section \ref{ssec:attrbidder}. We then explain in Section \ref{sec:metrics} how to measure the improvement offline with an updated version of the Expected Utility metric \cite{Chapelle2015}. We report and analyze positive results from both offline and online experiments in Section \ref{sec:experiments}. Finally we discuss future works and connections to other related approaches in Section \ref{sec:conclusion}.

\section{Model}

\subsection{Attribution Model}
\label{ssec:attrmodel}
The industry standard for conversion attribution is to credit the advertizing platform which owns the last click in a 30 days window before the conversion. Predicting the 30 days conversion probability is thus not enough as different events may change the attribution probability until the conversion actually happens: the competition (or other channels such as search-based advertising) may capture the attribution by generating clicks on their side. This characteristic of the last-click attribution has been shown to drive more efforts on average from advertising platforms and thus be more effective than other mechanisms in the case of cost-per-mil (CPM) payment models \cite{Berman2015}. However from the same study we observe that it is not obvious for performance advertising (CPC or CPA). It was also shown in \citep{Xu2016} that last-click attribution might prevent advertising platform to bid proportionally to the real marginal impact of the ad on the user conversion.

In the following model we do not explicitly assume that the advertiser uses last-click attribution to reward advertising platforms, but instead we propose to learn attribution from the data and then incorporate it in the bidder itself.  We consider the point of view of an advertising platform having access to advertiser conversions and their attribution label. We don't assume access to competition data such as clicks made by other platforms.  

Let us introduce the notation: $C$ is a  random variable representing the click, $S$ any conversion (attributed or not), $A$ an attributed conversion, all of which are Bernoulli. $X$ is the generic random variable representing all available contextual information (such as the user context, website information, advertising campaign). $\Delta$ is the delay between a conversion and a click observed prior to it. The task of attribution modeling consists in learning $P(A=1|S=1,X=x)$ the probability of attributed conversion given there is a conversion, in a certain context $x$. 

To be used for driving the bidding strategy the attribution model needs to capture important aspects of the attribution process such as the implementation used by the advertiser and the characteristics of the competition. We propose to use mainly two sources of information:
\begin{itemize}
\item $\delta$: the time since the last click generated by the advertising platform that captures the uncertainty of still having the attribution in the face of competitive events
\item $x$: capturing global characteristics such as per advertiser implementation, user activity level , etc.
\end{itemize} 

The proposed model is an exponential decay model of $\delta$: the time between the conversion and the last click, parametrized by a decay factor $\lambda$:
\begin{equation}
P(A = 1\vert S = 1, X=x, \Delta=\delta) = e^{- \lambda(x) \delta} ~,~ \lambda(x) \ge 0
\label{eq:attr_model_simple}
\end{equation}
The $\lambda$ decay factor represents here the speed at which the attribution probability decreases with time after the click (taking into account events such as competitors winning the attribution thanks to clicks on their ads). This model is not specific to last-click attribution. It may work under other attribution schemes as soon as the attribution probability vanishes over time after the clicks. 
The choice of this exponential decay is quite common in survival analysis \cite{Kalbfleisch2002} and was also used to model delays in conversion feedback \cite{Chapelle2014}. Models that better fit the data such as Kaplan-Meier or Weibull \cite{Kalbfleisch2002} could be considered, but fine tuning the attribution model is not in the scope of the paper.

\subsubsection{Model Estimation}
\label{sssec:modeloptim}
We estimate $\lambda$ by maximum likelihood. Let's consider a dataset built using clicks-conversion log as follows: $\{(\delta_i, a_i)\}_{i=1 \ldots n}$ where $\delta_i$ is the time since last click of that particular conversion and $a_i$ the attribution: 1 if attributed, 0 if not to the advertising platform. Under the assumption $P(a_i=1 ~|~ \delta_i) = e^{-\lambda \delta_i}$ we can write the conditional likelihood of the data and our convex cost function, the negative log-likelihood.
\begin{equation}
LH(\lambda) = \prod_{i=1}^n P(A_i=a_i ~|~ \delta_i) = \prod_{i=1}^n e^{-\lambda a_i \delta_i}(1-e^{-\lambda \delta_i})^{1-a_i}
\end{equation}
\begin{equation}
\label{eq:nllh}
NLLH(\lambda) = \sum_{i=1}^n \lambda a_i \delta_i - (1-a_i)log(1-e^{-\lambda \delta_i})
\end{equation}

\subsection{Attribution-aware Bidder}
\label{ssec:attrbidder}
\begin{figure}[!t]
\includegraphics[scale=0.5]{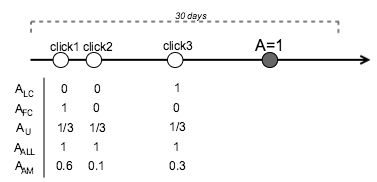}
\caption{Internal attribution procedure from the advertising platform point of view in a case where 3 clicks precede an attributed conversion. \textmd{5 possibles schemes are illustrated: last-click $A_{LC}$, first click $A_{FC}$, uniform $A_{U}$, all-clicks $A_{ALL}$ and an attribution equivalent to our attribution model $A_{AM}$.}}
\label{sales_matching_timeline}
\end{figure}

As mentioned before, the standard bidding strategy, in $2^{nd}$ price auctions,  is to bid the expected value of an impression \cite{Perlich2012}.  
As last-click attribution is usually assumed for the advertiser, previously published works also use last-click attribution when learning the conversion models, meaning that when several clicks occur before a conversion in the logs, only the last one is labelled as positive while previous ones are labelled as negative samples. Figure \ref{sales_matching_timeline} gives an illustrative example of the internal attribution procedure when 3 clicks occurred before an attributed conversion for a given advertising platform. Let us call $A_{LC}$ the random variable that measures if there was an attributed conversion in the 30 days after the click and this click was the last for the advertising platform. The bid value for an impression is estimated as:
\begin{equation}
bid \propto CPA \times P(A_{LC} = 1 \vert X=x)
\label{lc_bidder_equation}
\end{equation}

We can consider simple variants of this strategy by applying rule-based heuristics on the attribution process, by considering, for example, first-click ($A_{FC}$), uniform  ($A_{U}$) or linear attribution (some of them are depicted in in Figure \ref{sales_matching_timeline}). In fact we could argue that first clicks might have contributed more to user conversions than last ones. However, there is no evidence that fixed rules can work without making very strong hypotheses on the advertiser attribution scheme. Another alternative would be to consider all clicks leading to a conversion as positive examples ($A_{ALL}$) but, as other fixed rules, such a scheme cannot catch temporal dependencies between the auctions, which represents a key aspect in proper attribution modeling.  In the example of Figure \ref{sales_matching_timeline}, how should the conversion be attributed to each click? The first click might have more impact on the conversion but as the third one is closer to the conversion so it might also be important. In all cases click2 seems to have less importance than the other two. 

In line with lift-based bidding \citep{Xu2016} we propose to bid proportionally to the marginal attribution brought by the current opportunity: if the user recently clicked on one of our ads, we already have some probability to get the attribution and showing more ads  to the user might not be relevant for her given the recent interaction she had with the advertiser. 
To estimate the current attribution probability, we leverage the attribution model presented in Section \ref{ssec:attrmodel}.

At auction time, the gap between the current opportunity and the conversion $\delta_s$ if the user eventually converts is unknown. However,  we can measure the time elapsed since the last observed click $\delta_c$, which gives a good proxy for estimating the uncertainty under which we can hope to get an attribution from past auctions.
The attribution probability from the previous-click only is, according to Equation (\ref{eq:attr_model_simple}): $e^{-\lambda(x) (\delta_{s}+\delta_{c})}$. Now, if current auction leads to another click, our attribution probability becomes $e^{-\lambda(x) \delta_{s}}$. The marginal contribution of the  second click, normalized so that we can fully explain the conversion is then:
\begin{equation}
\frac{e^{-\lambda(x) \delta_{s}} - e^{-\lambda(x) (\delta_{s}+\delta_{c})}}{e^{-\lambda(x) \delta_{s}}} = 1-e^{-\lambda(x) \delta_{c}}
\end{equation}

The attribution-aware bidder directly modulates the attributed conversion probability by the marginal contribution of this opportunity:

\begin{equation}
b(x) \propto CPA \times P(A_{ALL}=1|X=x)(1-e^{-\lambda(x) \delta_{c}})
\label{aa_bidder_equation}
\end{equation}
where $\delta_{c}$ is the observed time since last click, and $A_{ALL}$ refers to whether the display led to an attributed conversion or not.  The first part of the model $P(A_{ALL}=1|X=x)$ represents the post-click conversion probability without any assumption on the internal attribution process, while second term handles the attribution part.

\subsubsection{Discussion}
If there is no previous click, second term vanishes and the bidder acts as the baseline EVB bidder, with one notable difference: all clicks leading to a conversion are labelled as positive, while only the last-one would be positive in the baseline.
The fact last-clicks of a given advertising platform are not favored in the learning is good from the advertiser perspective: the platform is not incentivised to "steal" the attribution when the true impact of the advertising on the conversion is low.
On the other hand, right after a click, the attribution bidder will drastically decrease the bid values. The recent click already gives a good probability of getting the attribution would there be a conversion. We also think that displaying new ads after a click does not increase the user propensity to convert, so investing the money elsewhere is a better strategy.  Ideally, as explained in \citep{Xu2016}, if the advertiser attribution process credits each event according to its causal impact on the conversion a bidder should only bid the lift in the conversion probability.

As a click is a strong engagement signal from the user we expect the law of diminishing return to apply to the impact of subsequent displays. However, baseline EVB models tend to bid higher after a click which increase both advertising costs and user exposure. The proposed bidder is thus closer to a lift-based bidder.

\section{Metrics}
\label{sec:metrics}

\subsection{Offline Metrics in Online Advertising}
Offline evaluation of bidding policies is especially important in advertising as online experiments imply to invest real money in ad auctions. Therefore offline evaluation consists in replaying logged auctions that were won by a previous production policy \cite{Zhang2014}. Lost auctions are not used as the bid needed to win the auction is unknown (which impairs counter-factual evaluation due to selection bias).
Still, as the state-of-the-art EVB strategy is to bid one's expected value i.e. $\text{payoff}  \times \text{conversion\_probability} $, classical offline metrics focus on the prediction problem as the payoff is usually fixed for a given campaign. Weighted Mean Squared Error \cite{Vasile2017} combine event payoff and prediction quality in a single metric. However such metric may not accurately predict online results if the competition changes. Recently \cite{Chapelle2015} proposed Utility to tackle this problem by injecting noise in the competing bids distribution. 

We now briefly describe the Utility metric before exploring its limits with respect to attribution in the following section.

Consider $v_i$, the value for the advertising platform of an action on an ad (e.g. a conversion value), $p_i$ the estimated probability of a user performing that action. In this setup the value of the impression can be estimated as $v_i \times p_i$. Let $c_i$ be the observed price, which is the highest competing bid in a second price auction, and $a_i$ a binary variable that represent if the action is attributed to the display or not. Then the payoff of the auction can be written as:
\begin{equation}
u_i = (a_i  \times v_i - c_i)\mathbbm{1}_{[p_i \times v_i >c_i]}
\end{equation}
The Utility of a new prediction model $p'$ is estimated on dataset $D$ as:
\begin{equation}
U(p') = \sum_{i \in D} (a_i  \times v_i - c_i) \mathbbm{1}_ {[p'_i v_i > c_i]}
\end{equation}

Expected utility was proposed to  model potential changes in the competing bids using a cost distribution $\Pr(c | c_i)$ instead of the actual observed cost:
\begin{equation}
EU(p') = \sum_{i \in D} \int_0^{p'_i \times v_i} (a_i  \times v_i - c) \Pr(c|c_i) dc
\end{equation}

A good choice for  $\Pr(c | c_i)$ is a Gamma distribution, parametrized by $\alpha, \beta$:  It interpolates between two extreme cases: empirical utility and WMSE \cite{Chapelle2015}:
\begin{equation}
\Pr(c|c_i) \sim \Gamma(\alpha=\beta c_i + 1, \beta)
\end{equation}
As in \cite{Chapelle2015} we use: $\alpha = \beta c + 1$.

\subsection{Attribution-aware Expected Utility}
\label{ssec:attrutility}
The expected utility basically supposes the conversions are attributed implicitly on a single display, usually the last-clicked display before the conversions. Evaluating other bidders on this metric will naturally favor last-click based bidder. Therefore, we propose to extend the metric by taking into account the attribution evolution. For this purpose we weigh the value of each display to make it depend on the attribution function $a: x \in \mathbb{R}^N \mapsto \mathbb{R}$:
\begin{equation}
\label{eq:attr_utility}
AEU(p',a) = \sum_{i \in D} \int_0^{p'_i \times v_i} (a(x_i) \times v_i - c) \Pr(c|c_i) dc
\end{equation}

If the attribution rule is last-click, then AEU is exactly equivalent to Expected Utility:
\begin{equation}
    a_{lc}(x)=
    \begin{cases}
      1 & \text{if}\ x\ \text{is last-click} \\
      0 & \text{otherwise}
    \end{cases}
  \end{equation}
  
A natural choice for the attribution function is to re-use the attribution model proposed in Section \ref{ssec:attrmodel}: 
\begin{equation}
    a_{am}(x)=
    \begin{cases}
      1 - e^{- \lambda(x) \delta_c} & \text{if}\ x\ \text{was clicked} \\
      0 & \text{otherwise}
    \end{cases}
\end{equation}
As this attribution scheme is not normalized, each conversion may not contribute equally to the metric (attributions of all displays preceding a conversion might not sum to one). This tends to penalize conversions for which the last click occurred long time ago: we believe it is a desired property for a metric to limit the importance of conversions that occurred days or even weeks after clicks,  as the real causal impact of these clicks is expected to be small. Anyway we also report in the experiments a normalized version (where each conversion contributes the same amount, as shown by $A_{AM}$ in Figure \ref{sales_matching_timeline}) in addition to the raw one.

A risk of the $AEU$ is to choose a bidder that overfits the learned attribution model. In that case we advise to consider a regularized version that, for example,  would not be strictly dependent on the lambda parameter of the model. It is however not in the scope of this paper and will be left for future research.

In the remaining of the paper, for sake of simplicity,  we call $U_A$ the $AEU$ using normalized proposed attribution model, $U_A*$ its un-normalized counterpart and  $U_{LC}$ the variant with last-click attribution function.

\section{Experiments}
\label{sec:experiments}

\subsection{Criteo Attribution Dataset}
Attribution information (the difference between all sales observed on an advertiser site and attributed ones) is mandatory for our study and not present in previously published conversion datasets \cite{Criteo2014} \cite{Zhang2014}. Therefore for all offline experiments we use a log sampled from 30 days of Criteo production traffic with displays, clicks, conversions and attribution data. It will be released publicly at \url{http://research.criteo.com}. The data has been sub-sampled and anonymized so as not to disclose proprietary elements but the exact method is irrelevant here. Each line represents an ad impression with following information: timestamp, price paid, categorical features of the user, ad and publisher, click*, position of the click*, conversion*, conversion value*, attribution*.


Elements marked with a * are present if applicable. Conversion and their value are credited by the advertiser and made comparable across the dataset. If multiple clicks occurred and there was a conversion the click position is provided, which is necessary for implementing different attribution schemes.

\subsection{Attribution Study}

\label{ssec:attrib_study}
\begin{figure}[!t]
\includegraphics[scale=0.5]{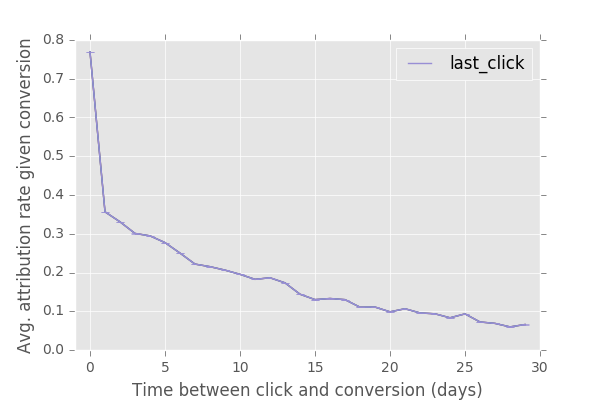}
\caption{Average attribution given conversion and time since last click}
\label{avg_attribution_given_tslc_and_conversion}
\end{figure}

The model presented in Section \ref{ssec:attrmodel} encodes the attribution probability as a function of the time between a click and the conversion. The empirical attribution rate as function of this delay is illustrated on Figure\ref{avg_attribution_given_tslc_and_conversion} and shows a sharp decline in attribution rate after a click followed by a slower decrease tending to $0$ after 30 days.

\begin{figure}[!t]
\includegraphics[scale=0.5]{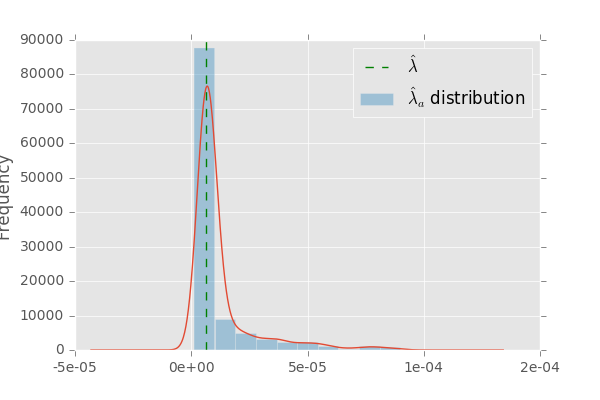}
\caption{Distribution of per advertiser $\hat{\lambda_a}$ versus global $\hat{\lambda}$}
\label{lambda_per_advertiser}
\end{figure}

We estimated a global $\hat{\lambda} \approx 6.25 \times 10^{-6}$ on the log. It is remarkably stable over time with a daily relative variation $< 0.1\%$.  
As seen on Figure \ref{lambda_per_advertiser} the $\lambda$ parameter is also stable across advertisers. We posit that fitting one parameter per advertiser would only improve the training objective marginally and thus prefer to use a global one.

\subsection{Offline Evaluation}

\subsubsection{Baselines}

\begin{figure}[!t]
\includegraphics[scale=0.5]{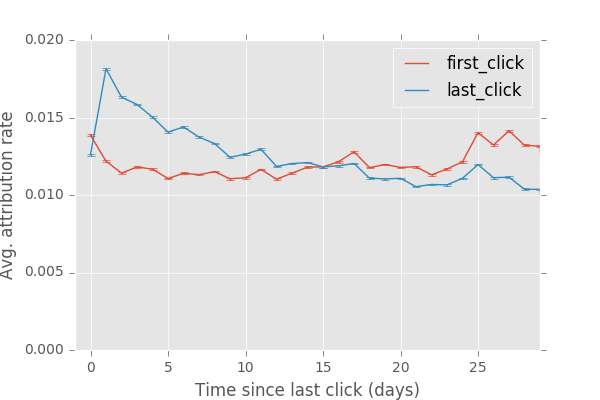}
\caption{Average attribution rate per display, given time since last click for two attribution schemes}
\label{avg_attribution_given_tslc}
\end{figure}

The tested bidding strategies include two expected value bidders. LCB is the traditional baseline defined in Eq. \ref{lc_bidder_equation}. First-click bidder (FCB) is its first-click counterpart. Both actually use the same features but different labels for learning the conversion model. To understand the difference we observe the attribution rate with respect to an important variable: time since last click, as illustrated on Figure \ref{avg_attribution_given_tslc}. We observe a high proportion of positive examples for LCB within a short delay after a click and that proportion decreases with the delay. Conversely, for FCB the initial proportion is low and increases slowly over time. 

Our proposed attribution bidder (AB) defined per Equation \ref{aa_bidder_equation} is composed of two multiplicative parts. As described earlier the bid values will drastically decrease right after clicks as can be deduced from Figure \ref{avg_attribution_given_tslc_and_conversion}.

\subsubsection{Evaluation Setup}
Our evaluation protocol is as follows: the last 7 days of log are considered for testing and for each test day we use the 21 previous days for learning.
For learning conversion models, all the variables are mapped into a sparse binary feature vector via the hashing trick and an $L^2$-penalized logistic regression model is trained using the L-BFGS algorithm (similar to  \cite{Chapelle2014a}).
The evaluation metrics are variants of the Attribution-aware Expected Utility: $U_A$, $U_A*$, and $U_{LC}$ as presented in Section \ref{ssec:attrib_study}. In order to have comparable values between bidders for a given metric, predictions are calibrated so as to predict the same value on average. In a real system this calibration property is usually controlled by a feedback loop mechanism to ensure constant spend. More details are given in Section \ref{ssec:online_exp}.
For each variant of Utility, we use different $\beta \in (1000/inf)$ which correspond to medium/no variations of the competing bids.

\subsubsection{Results \& Discussion}

\begin{table}[t!]
\centering

\begin{tabular}{lcc}
{} & {$U_{A}$} &  {$U_{LC}$}\\
\midrule
AB vs LCB & +12.32\% & -14.15\% \\
\bottomrule
\end{tabular}

   \bigskip
	\captionsetup{justification=raggedright,margin=.5cm,skip=3pt}
	\caption{
	Offline comparison of Attribution (AB) versus Last-Click (LCB) bidders.
	\textmd{$U_A$ and $U_{LC}$ refer to variants of $AEU$ with resp. last-click or proposed attribution model and cost perturbation $\beta=1000$. Result is significant at the $.05$ quantile of 100 bootstraps resamples.}
	}
	\label{tab:delta_attrib_lc}
\end{table}

\begin{table}[t!]
\centering
\begin{tabular}{lrrr}
\rule{0pt}{4ex}  &                  LCB &                 FCB &                 AB \\
\midrule
Win Rate                      &                        0.94 &                        0.90 &                        0.89 \\
\midrule
$U_A, \beta_{1000}$         &               $1679 \pm 31$ &               $1843 \pm 37$ &  $1886 \pm 39$ \\
$U_A, \beta_{inf}$          &               $3383 \pm 46$ &               $3471 \pm 52$ &  $3549 \pm 58$ \\
\midrule
$U_A*, \beta_{1000}$        &               $2852 \pm 43$ &               $2888 \pm 43$ &  $\boldsymbol{3396} \pm 53$ \\
$U_A*, \beta_{inf}$         &               $5105 \pm 56$ &               $5083 \pm 57$ &  $\boldsymbol{5408} \pm 61$ \\
\bottomrule
\end{tabular}
    \bigskip
	\captionsetup{justification=raggedright,margin=.5cm,skip=3pt}
	\caption{
	Offline comparison of 3 bidders (Last-click LCB, first-click (FCB) and attribution bidder (AB)).
	\textmd{$U_A$/$U_A*$ refers to Eq. \ref{eq:attr_utility} with cost perturbation $\beta$ and normed/un-normed attribution model.}
	}
	\label{tab:offline_results}
\end{table}

%

Our first result is summarized in Table \ref{tab:delta_attrib_lc}: as expected the proposed AB is favored by the  Utility when the attribution is given by the same attribution model. Conversely, LCB is favored by the last-click Utility. We notice in this result how LCB becomes sub-optimal when the metric focuses on earlier clicks. At this point we posit that the Utility that uses an attribution model $U_A$ best reflects the real attribution system.

Our second result is summarized in Table \ref{tab:offline_results}: when measured across different variants of the  $U_A$, the proposed AB policy performs the best among the three. The reported win rate tells us that LCB wins 5\% more auctions than AB, auctions that are probably less valued by the metric, representing excessive post-click user exposure.
Event though $FCB$ performs quite on par with $AB$ on the normed Utility $U_A$ this is not the case with the un-normed $U_A*$ version and the gap grows larger with smaller $\beta$ ($\beta_{inf}$: +6\% / $\beta_{1000}$:+17\%), i.e when there is more uncertainty on the competing bids. This let us think AB is more robust in real systems. This confirms the comments in Section \ref{ssec:attrutility}.
Thoroughly evaluating $U_A$ deserves to study the correlation of offline versus online results. We deliberately skip such analyses in the current work as it is not the main focus of this paper.

\begin{figure}[!ht]
\includegraphics[scale=0.5]{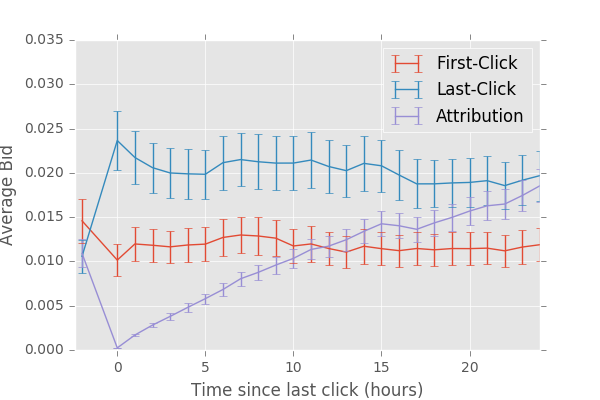}
\caption{Average bid given time since last click for different bidding strategies}
\label{emp_bid_profiles}
\end{figure}

Our third result presented in Figure \ref{emp_bid_profiles} illustrates the average post-click bid evolution for the different bidders in the first 24 hours. As expected, LCB overbids strongly just after a click and then slowly converges to the nominal pre-click bid. Conversely, FCB underbids after a click (although less sharply than LCB) and overbids increasingly compared to the pre-click level. We observe that our proposed AB policy bids at an extremely low level just after a click and then steadily increases its bids.

\subsection{Online Experiments}
\label{ssec:online_exp}
\subsubsection{Experimental Setting}
We fit an attribution model with a global $\lambda$ using the procedure described in Section \ref{sssec:modeloptim}. We learn an attribution model encompassing the business standard which means looking back up to 30 days before the conversion for potential clicks.

Our online experimental policy is a simple multiplier applied to all bids:
\begin{equation}
\label{bid_factor}
bid_{test} \propto bid_{ref} ~ A ~ (1 - B ~ e^{-\lambda \delta})
\end{equation}
This implementation is straightforward and flexible for different practical use cases. Firstly, it allows for more or less aggressive bid decreases thanks to the $B$ factor: small $B$ will only slightly change the reference bidding policy. Secondly, the proposed bidding policy consists only in reducing bids and hence should result in under-spending with respect to the reference population. To compare policies one should be able to measure their performance in a setting where both spend the entire available advertiser budget. The $A$ factor can be used to equalize spent budget between the control and test populations for this purpose. 

Different values for the $A$ and $B$ parameters have been A/B tested by user split on the Criteo production platform.

\subsubsection{Results \& Discussion}

\begin{table}[ht!]
  \centering
  \begin{tabular}{cccc}
  \toprule
    \begin{tabular}{@{}c@{}}$\Delta OEC$ \\(long term)\\ \end{tabular} & \begin{tabular}{@{}c@{}}Revenue \\(short term)\\ \end{tabular} & \begin{tabular}{@{}c@{}}Advertiser \\ROI\\ \end{tabular} & \begin{tabular}{@{}c@{}}User ad \\exposure\\ \end{tabular}\\
    \midrule
    +5.5\% & negative & positive & lower\\
    \bottomrule
  \end{tabular}
  \bigskip
  \captionsetup{justification=raggedright,margin=.5cm,skip=3pt}
  \caption{Online results summary. \textmd{\small{OEC is a long-term performance metric. We report the uplift of the proposed bidder wrt. the baseline production bidder. All reported results are significant at the .05 quantile of 100 bootstraps resamples.}}}
  \small
  
  \label{tab:online_results}
\end{table}

Overall results are summarized in Table \ref{tab:online_results}. We found that the proposed bidding policy showed positive long-term results in all settings. The Overall Evaluation Criterion (OEC) \cite{Kohavi2007} reported for this experiment captures the value generated by Criteo for the advertisers: $OEC \propto (AdvertiserValue - Cost)$ and is not necessarily correlated to short-term revenue. The consistent improvement of the OEC by +5.5\% across all production traffic is remarkable since it is quite rare given the difficulty to move such long term criteria.

The OEC was moved mainly through reducing bids and hence budget spend in subsequent auctions. The accompanying negative impact in revenue shall be explained as follows. A better efficiency of the proposed bidding policy delivers more value for less cost, hence reducing the spend of advertisers (and Criteo revenue) to achieve a target CPA. Of course rational advertisers should then increase their target CPA to benefit from additional value. 

From a user perspective such policies reduced drastically the exposition to ads from a given advertiser after a click on an ad of that advertiser. Incidentally, and beyond overall reduced exposure, it brought more diversity on ads seen as other advertisers on the Criteo platform had now better chances of winning bids.

For advertisers we noticed an increased return on investment (ROI) as measured by the number of conversions (or conversions values) generated for a given budget.

\section{Conclusion}
\label{sec:conclusion}

We proposed a novel, effective yet simple bidding policy leveraging attribution modeling. Our model can be trained efficiently and requires minimal changes to be implemented in an advertising platform. We proposed a modification of the Utility offline metric. We also reported highly positive experiments both offline and online at scale on Criteo production platform.

We plan to release the dataset used for offline experiments under the same conditions as previous Criteo datasets so as to foster additional research in attribution modeling. An immediate perspective would be to study alternative attribution models and their impact. Future works could also include researching this problem in i) the reinforcement learning setting where the state would incorporate the attribution evolution and ii) the online learning framework where attributed value is learnt across repeated auctions in the spirit of \cite{Weed2015}.

\section*{Acknowledgments}
The authors would like to thank the Criteo team and especially Nicolas Le Roux, Clément Mennesson, Alexis Watine for helpful discussions. 

\bibliographystyle{apalike}
\bibliography{paper}
\end{document}